\documentclass[letterpaper]{article} 
\usepackage{aaai2027}  
\nocopyright

\usepackage[hyphens]{url}  
\usepackage{graphicx} 
\usepackage{natbib}  
\usepackage{caption} 
\usepackage{algorithm}
\usepackage{algorithmic}
\usepackage{amsmath}
\usepackage{amssymb}
\usepackage{xspace}

\newcommand{\name}{\textit{IDraw}\xspace}

\usepackage{newfloat}
\usepackage{listings}
\DeclareCaptionStyle{ruled}{labelfont=normalfont,labelsep=colon,strut=off} 
\floatstyle{ruled}
\newfloat{listing}{tb}{lst}{}
\floatname{listing}{Listing}

\usepackage{booktabs}
\usepackage{array}
\usepackage{multirow}
\usepackage{dblfloatfix}
\usepackage{cuted}
\usepackage{pifont}
\usepackage{subcaption}

\title{\textit{IDraw}: Artist Verification from Digital Drawing Images}

\author{
    Nayoung Kim\textsuperscript{\rm 1},
    Nan Jiang\textsuperscript{\rm 2},
    Bangjie Sun\textsuperscript{\rm 2},
    Jaewon Shin\textsuperscript{\rm 1},
    Sojeong Kim\textsuperscript{\rm 1},
    Jun Han\textsuperscript{\rm 1}\corresponding
}

\affiliations{
    \textsuperscript{\rm 1}Korea Advanced Institute of Science and Technology (KAIST)\\
    \textsuperscript{\rm 2}National University of Singapore (NUS)\\
    skdud727@kaist.ac.kr,
    jiangnan@u.nus.edu,
    bangjie@nus.edu.sg,\\
    sjw0903zz@kaist.ac.kr,
    cindy0325@hanyang.ac.kr,
    junhan@cyphy.kaist.ac.kr
}

\begin{document}

\maketitle

\begin{abstract}
As digital drawings are increasingly shared online, reliable authorship verification has become important for protecting artists and resolving disputes. Yet when authorship is questioned, verification may have to rely only on the disputed drawing and reference drawings known to be created by the claimed artist. 
This setting is challenging for two reasons. First, artist-specific drawing behavior, such as pen pressure and movement speed, is informative but is not available from a completed drawing. Second, similarities in the depicted object or scene can obscure similarities arising from the artist. 
We propose \name, a framework that learns from drawings paired with tablet-pen sensor signals collected from separate training artists. This allows \name to infer drawing behavior from completed images during a later authorship dispute, without requiring sensor data from the artist being verified. \name also reduces the influence of drawing content by identifying information shared by drawings of the same object across different artists and suppressing it before comparing drawings. 
To support this approach, we construct the first multimodal dataset for digital drawing authorship verification, containing 1,110 drawings from 37 artists and 14 types of tablet-pen sensor signals. Evaluated on previously unseen artists across nine image-encoder backbones, \name consistently outperforms standard image-based verification and reduces verification error by up to 40\%. These results demonstrate that inferring drawing behavior from completed images and suppressing drawing content improve digital drawing authorship verification.

\end{abstract}


\section{Introduction}
\label{sec:introduction}

As digital drawings are increasingly produced and shared online, disputes over who actually created a drawing are becoming a recurring concern~\cite{fiesler:15,mcconaghy:17}. Artists may be miscredited, drawings may be falsely claimed by others, and works may be published under someone else’s name. In such cases, verification may have to rely only on the disputed drawing and other drawings known to be created by the claimed artist. This motivates digital drawing authorship verification: determining whether the disputed drawing was created by the claimed artist.

However, verifying an artist from a completed drawing is far from trivial because informative drawing-process signals are no longer available. When a drawing is produced using a tablet and digital pen, signals such as pen pressure and movement speed can capture artist-specific drawing behavior. Behavior-based methods use such signals to distinguish individuals, but require access to the drawing process~\cite{muramatsu:07,tolosana:21}. Prior methods attempt to infer drawing behavior from completed images, yet recover only partial information, such as visible trajectories~\cite{li:21,qiao:07}. Hence, a key challenge is to infer artist-specific drawing behavior from the completed image without requiring sensor data from the artist being verified.

A second challenge is that the completed image reflects both the artist and the content being drawn. Artist-specific visual cues, such as line quality, shape construction, and shading patterns, are expressed through a particular object or scene. Hence, drawings of different objects by the same artist may appear dissimilar, while drawings of the same object by different artists may appear similar. Image-based methods operate directly on completed drawings, but their comparisons may reflect similarities in the depicted content rather than recurring characteristics of the artist~\cite{elgammal:18}. Hence, reliable authorship verification requires separating artist identity from drawing content.

\begin{figure*}[!t]
    \centering
    \includegraphics[width=\textwidth]{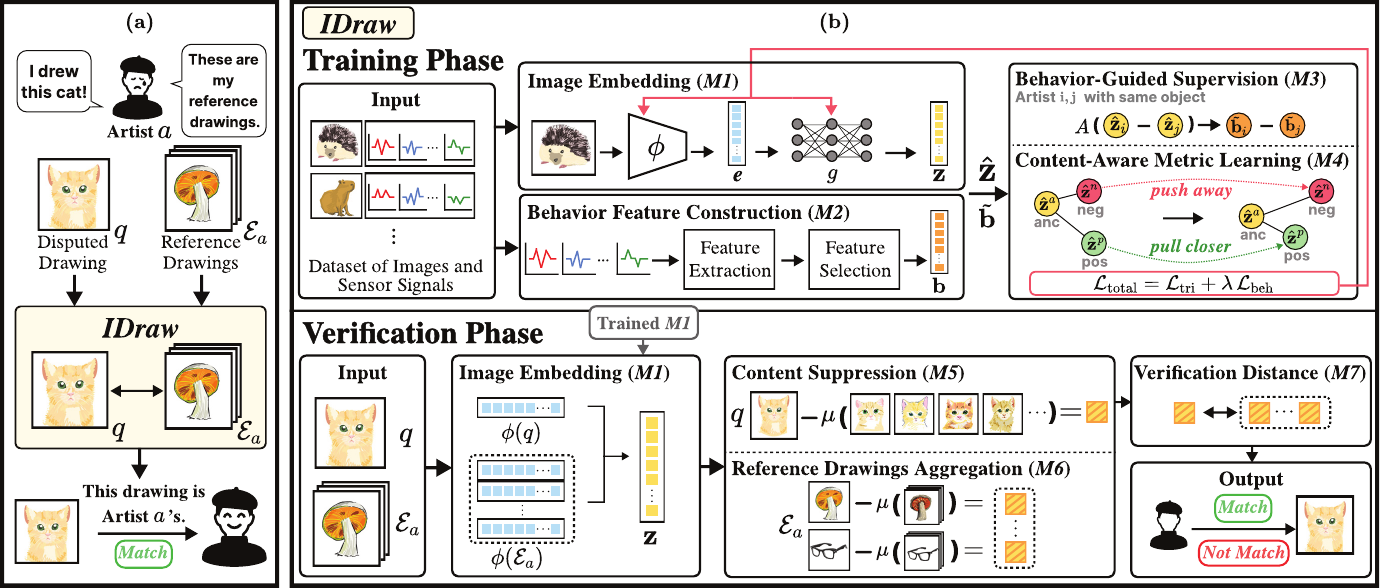}
    \caption{\textbf{Overview of \name.}
    \textbf{(a)} \name verifies whether a disputed drawing is created by the claimed artist using the artist's reference drawings. \textbf{(b) Training phase:} Modules \textit{M1} through \textit{M4} learn an image embedding guided by tablet-pen sensor signals and organized by artist identity. \textbf{Verification phase:} Modules \textit{M5} through \textit{M7} use only completed drawings to suppress remaining drawing content, aggregate the reference drawings, and compute the cosine distance to the query drawing.}
    \label{fig:design-overview}
\end{figure*}

\newcommand{\yes}{\checkmark}
\newcommand{\no}{--}
\begin{table*}[!t]
\centering
\footnotesize
\setlength{\tabcolsep}{2pt}
\begin{tabular*}{\textwidth}{@{\extracolsep{\fill}}l c r r c l@{}}
\toprule
\textbf{Dataset} & \textbf{Medium} & \textbf{Objects} & \textbf{Artists}
& \shortstack{\textbf{Fully}\\\textbf{crossed}}
& \textbf{Tablet-pen sensor signals} \\
\midrule
Painter by Numbers~\cite{painterbynumbers:16}
  & Physical & \no & 1{,}584 & \no & None (images only) \\
\addlinespace[1pt]
QuickDraw~\cite{quickdraw:16}
  & Digital & 345 & \no & \no & Timing \\
FS-COCO~\cite{fscoco:22}
  & Digital & 10{,}000 & 100 & \no & Timing \\
OpenSketch~\cite{opensketch:19}
  & Digital & 12 & 15 & \no & Timing, pressure \\
DifferSketching~\cite{differ:22}
  & Digital & 136 & 108 & \no & Timing, pressure \\
\midrule
\textbf{\name}
  & \textbf{Digital} & \textbf{30} & \textbf{37} & \textbf{\yes}
  & \textbf{Geometry, strokes, editing, timing, pressure, tilt} \\
\bottomrule
\end{tabular*}
\caption{\textbf{Comparison of datasets for drawing authorship and process.} \name is
the only one in which every artist draws every object, with tablet-pen sensor signals
recorded throughout creation.}
\label{tab:datasets}
\end{table*}
To address these limitations, we propose \name, a framework that verifies a \textit{disputed drawing} against a set of \textit{reference drawings}, i.e., drawings known to have been created by the claimed artist. As illustrated in Figure~\ref{fig:design-overview}(a), \name compares the disputed drawing with these reference drawings to determine whether they share the same authorship. \name is trained on drawings paired with tablet-pen sensor signals from a \textit{separate set of artists}. Using these paired data, it learns to \textbf{\textit{infer drawing behavior}} captured by signals such as pen pressure and movement speed from completed images, guiding the image encoder to capture \textit{behavior-related artist cues}. Once trained, the image encoder can capture these cues from both the disputed and reference drawings \textit{without requiring sensor data} from the artist being verified.
To reduce content dependence, \name also estimates the component shared by drawings of the same object across different artists and \textit{suppresses} it from each drawing's representation. The resulting \textbf{\textit{content-suppressed representation}} groups drawings by who created them rather than by what they depict. During verification, \name aggregates the content-suppressed representations of the reference drawings and compares the aggregated representation with that of the disputed drawing to determine authorship.

Supporting this training strategy requires drawings paired with sensor signals recorded during their creation. However, existing art and digital-drawing datasets either contain only completed images without drawing-process information~\cite{painterbynumbers:16,wikiart,omniart:18,mensink:14} or provide only a limited subset of tablet-pen signals~\cite{opensketch:19,differ:22}. Hence, we construct the first multimodal dataset for digital drawing authorship verification, containing 1,110 drawings from 37 artists, each of whom draws the same 30 objects. Every drawing is paired with 14 types of tablet-pen sensor signals, including pen pressure and movement speed. This dataset enables \name to learn how artist-specific drawing behavior is reflected in completed images.

We evaluate \name on previously unseen artists under a challenging setting in which the disputed drawing and the claimed artist’s reference drawings depict different objects. Across nine image-encoder backbones, \name consistently outperforms standard image-based verification and reduces verification error by up to 40\% relative to the baseline. Ablation studies further show that behavior-guided training and content suppression each contribute to these gains. These results demonstrate that \name can verify artists across changes in drawing content using only the disputed and reference drawings.

In summary, this work makes the following contributions:
\begin{itemize}

    \item We propose a behavior-guided training strategy that learns to infer \textit{drawing behavior} from completed images using tablet-pen sensor signals collected from separate training artists, while requiring only the disputed and reference drawings during verification.
    
    \item We introduce a content-suppression approach that estimates and suppresses information shared by drawings of the same object across different artists. The resulting \textit{content-suppressed representation} groups drawings by who created them rather than by what they depict.
    
    \item We construct the first multimodal dataset for digital drawing authorship verification, containing 1,110 drawings from 37 artists, each paired with 14 types of tablet-pen sensor signals recorded during creation.
    
    \item We evaluate \name on previously unseen artists across nine image-encoder backbones and demonstrate consistent improvements over standard image-based verification.

\end{itemize}

\section{Related Works}
\label{sec:related}

\subsection{Drawing Authorship Verification}
\label{sec:related-verficiation}
%
\paragraph{Image-based methods.}
Image-based methods infer authorship from the completed drawing. Authorship can be established proactively through watermarks~\cite{stegastamp:20} or provenance metadata~\cite{c2pa, camprints:25}, but these signals may not survive redistribution~\cite{regen:24}. We instead consider image-only verification, where no external mark remains.

On digitized paintings, convolutional networks classify
artists~\cite{tan:16,saleh:16,cetinic:18}, while stroke-level
properties such as thickness support finer-grained
attribution~\cite{elgammal:18}. These methods can be applied directly to
digital drawings, but their representations may be dominated by depicted
content. As a result, drawings may be grouped by what they depict rather
than by who created them. Verification methods in other domains reduce such
shared variation using statistics estimated from training
subjects~\cite{auckenthaler:00,christlein:17}. Our content suppression
adapts this idea to what same-object drawings have in common.

\paragraph{Sensor-based methods.}
Sensor-based methods identify creators from signals recorded during
creation. Tablet-pen sensor signals capture pressure, pen posture, stroke timing, trajectory, and editing behavior. Online signature verification methods use such signals to verify identity ~\cite{muramatsu:07,fierrez:07,diaz:19,lai:22,tolosana:21}. However, these methods cannot be used when only completed images are available.

Our work brings image and sensor information together, following cross-modal sensing work that learns human input behavior from complementary observations~\cite{unikey:25,keynergy:21,perceptio:18}. During training, tablet-pen sensor signals guide the image encoder to infer drawing behavior from completed images, while verification requires only the completed drawings.  Our setting further follows learning using privileged information, because sensor signals are available during training but not during verification~\cite{vapnik:09,lopezpaz:16,fedcmd:24,owens:16,lyu:24}.

\subsection{Datasets for Drawing Authorship and Process}
\label{sec:related-datasets}
Behavior-guided training requires completed drawings paired with the tablet-pen sensor signals recorded during their creation. Table~\ref{tab:datasets} compares existing datasets under this requirement. Large-scale artwork datasets collect digitized works, often with artist labels~\cite{painterbynumbers:16,wikiart,omniart:18,mensink:14}, although some lack artist annotations altogether~\cite{bam:17}. However, they preserve only completed images, not the creation process. Digital sketch datasets capture part of the drawing process, but provide only limited process data~\cite{quickdraw:16,fscoco:22,eitz:12,sangkloy:16, didi:20,opensketch:19,differ:22}. Neither type of dataset uses a fully crossed artist--object design in which every artist draws every object. This makes it difficult to separate differences between artists from differences between objects. Hence, we collect a fully crossed drawing--sensor dataset.

\section{Problem Formulation}
\label{sec:problem}
%
\paragraph{System Goal.}
The goal of \name is to determine whether a disputed drawing is created by a claimed artist. The input consists of the disputed drawing, which we call the query drawing $q$, a claimed artist $a$, and a set of reference drawings $\mathcal{E}_a$. All drawings in $\mathcal{E}_a$ are known to have been created by $a$. The system determines whether $a$ also created $q$.

\paragraph{Assumptions and Constraints.}
We assume that $\mathcal{E}_a$ is registered through an identity-verified channel before any dispute arises. Attacks on this registration process are outside the scope of this work. The query drawing $q$ is not included in $\mathcal{E}_a$. We consider a setting in which $\mathcal{E}_a$ excludes $q$ and contains no drawing of the same object as $q$. 

\paragraph{Threat Model.}
The \textbf{attacker's goal} is to make a false authorship claim, either by
presenting an existing artist's drawing as their own or by falsely claiming that
a particular artist created a drawing made by someone else. In terms of
\textbf{attacker's capability}, we assume that the attacker can obtain and
republish existing drawings or submit an arbitrary drawing with an authorship
claim through an account, caption, or metadata. The attacker may also create or
select drawings with the same or similar content as the claimed artist's work.
During verification, the disputed drawing is compared with the trusted reference
drawings of the claimed artist. \name then determines whether the authorship
claim is valid.

\section{Methodology}
\label{sec:design}

\subsection{Overview}
\label{sec:design-overview}
We propose \name, an image-only framework for digital drawing authorship verification. This task poses two challenges: (1) behavior-related artist cues are not directly observable from a completed image, and (2) drawing content can dominate the image embedding. To address these challenges, we use tablet-pen sensor signals only during training to guide the image embedding.

Figure~\ref{fig:design-overview}(b) presents the framework. The training phase has four modules. \textit{M1} maps each completed drawing to an image embedding. \textit{M2} converts its paired tablet-pen sensor signals into a fixed-size drawing behavior feature vector. \textit{M3} uses pairs of same-object drawings to guide the normalized image embedding to encode behavior-related artist cues. In addition, \textit{M4} places same-artist drawings close across different objects and same-object drawings by different artists far apart, reducing the influence of drawing content.

In the verification phase, \name receives a query drawing and the claimed artist's reference drawings. \textit{M5} explicitly suppresses the remaining drawing content by subtracting the corresponding object mean from each embedding. \textit{M6} averages the content-suppressed reference representations to obtain the reference artist representation. \textit{M7} computes the cosine distance between the content-suppressed query representation and the reference artist representation as the verification distance.

\begin{figure}[!t]
    \centering
    \includegraphics[width=0.9\columnwidth]{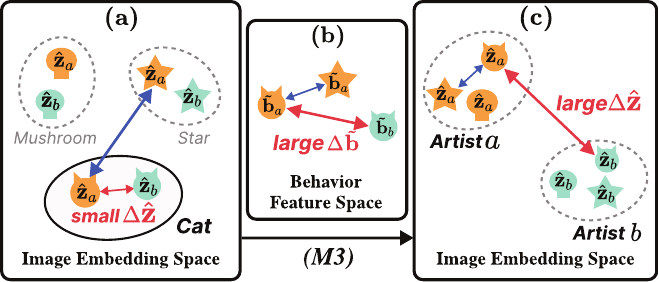}
    \caption{\textbf{Behavior-guided supervision (\textit{M3}).}
    \textbf{(a)} A pair of same-object drawings by different artists has a small image
    embedding difference.
    \textbf{(b)} The same drawing pair has a large difference in drawing behavior.
    \textbf{(c)} \textit{M3} makes the image embedding difference
    ($\Delta\hat{\mathbf{z}}$) reflect the drawing behavior difference
    ($\Delta\tilde{\mathbf{b}}$).}
    \label{fig:design-module}
\end{figure}
%
\subsection{Training Phase}
\label{sec:design-training}
%
\paragraph{Image Embedding (\textit{M1}).}
The completed drawing and its sensor signals are different modalities, so we represent both as vectors to relate them during training. \textit{M1} maps each completed drawing to a 256-dimensional \textbf{image embedding} used by the subsequent modules.

Given a digital drawing $x$, a trainable image encoder $\phi$ produces an image feature $\mathbf{e}=\phi(x)$. We standardize each dimension of $\mathbf{e}$ using training-set statistics to obtain $\tilde{\mathbf{e}}$. A trainable projection head $g$ then maps this feature to the image embedding
%
\begin{equation}
    \mathbf{z}
    =
    g\left(\tilde{\mathbf{e}}\right)
    =
    W\tilde{\mathbf{e}}
    +
    h\left(\tilde{\mathbf{e}}\right)
    \in
    \mathbb{R}^{256},
    \label{eq:embedding}
\end{equation}
%
where $W$ is a linear projection and $h$ is a two-layer multilayer perceptron. The encoder $\phi$ and projection head $g$ are trained jointly.

We standardize $\mathbf{z}$ per dimension using the training-set mean and
standard deviation, yielding $\mathbf{u}$. We then apply $\ell_2$ normalization to obtain $\hat{\mathbf{z}}$. Both training objectives operate on $\hat{\mathbf{z}}$.

\paragraph{Behavior Feature Construction (\textit{M2}).}
A completed image and its raw sensor signals have no direct temporal alignment. The image preserves only partial traces of the drawing process, while sensor signals vary in length and cannot be aligned across drawings. Hence, \textit{M2} summarizes each sequence as a fixed-size drawing behavior feature vector that captures overall drawing patterns rather than raw sensor signals. This vector can then be related to the image embedding during training.

Following prior work on pen-based behavior analysis~\cite{fierrez:07,emothaw:17,drotar:16}, we first extract 109 candidate features from the raw sensor signals, such as average pen pressure and mean stroke curvature. We then select the behavior features in two steps. First, we discard features that can be computed directly from the completed drawing, such as color usage and object shape, leaving 64 features. Second, we remove near-duplicate features with $|\rho_{\mathrm{Spearman}}| \ge 0.95$. We also exclude nine pen-tilt features because our interface does not render pen tilt in the completed drawing.

This process produces a $K$-dimensional \textbf{drawing behavior feature vector} $\mathbf{b}\in\mathbb{R}^{K}$, where $K=50$. We standardize each dimension using training-set statistics to obtain $\tilde{\mathbf{b}}$. The selected features cover stroke geometry, pressure, timing, and editing behavior. Their definitions, physical interpretations, and quality-control procedure are provided in the supplementary material.

\paragraph{Behavior-Guided Supervision (\textit{M3}).}
Drawing behavior provides artist-specific cues, but sensor signals are unavailable during image-only verification. Hence, \textit{M3} encourages the normalized image embedding to encode behavior-related artist cues. We use pairwise supervision because verification compares drawings rather than evaluating each drawing independently. Pairing drawings of the same object holds the depicted object fixed, allowing the supervision to focus on differences in how the artists produced them.

Figure~\ref{fig:design-module} shows that two same-object drawings by
different artists can have a small difference between their normalized image
embeddings in (a), but a large difference between their standardized drawing
behavior feature vectors in (b). For two drawings, $\Delta\hat{\mathbf{z}}$ denotes the difference between their normalized image embeddings, and
$\Delta\tilde{\mathbf{b}}$ denotes the difference between their standardized
drawing behavior feature vectors. \textit{M3} encourages the embedding
difference to reflect the large behavior difference, making the embedding
more sensitive to behavior-related artist cues, as illustrated in Figure~\ref{fig:design-module}(c).


A linear head $A\in\mathbb{R}^{K\times256}$ predicts $\Delta\tilde{\mathbf{b}}$ from $\Delta\hat{\mathbf{z}}$ by transforming $\Delta\hat{\mathbf{z}}$ to match the dimension and scale of $\Delta\tilde{\mathbf{b}}$.  We define the \textbf{behavior loss }as the mean squared error for this prediction:
\begin{equation}
    \mathcal{L}_{\mathrm{beh}}
    =
    \frac{1}{K}
    \left\|
    A\Delta\hat{\mathbf{z}}
    -
    \Delta\tilde{\mathbf{b}}
    \right\|_2^{2}.
    \label{eq:behavior}
\end{equation}
The loss is averaged over same-object pairs in each batch.

\paragraph{Content-Aware Metric Learning (\textit{M4}).}
\textit{M3} encourages the image embedding to encode behavior-related artist cues, but it does not determine which drawings should be close or far apart. Verification compares drawings by their embedding distance. Hence, \textit{M4} places drawings by the same artist close together even when they depict different objects, and drawings by different artists far apart even when they depict the same object.

For an anchor $\hat{\mathbf{z}}^{a}$, the positive $\hat{\mathbf{z}}^{p}$ is a different-object drawing by the same artist, and the negative $\hat{\mathbf{z}}^{n}$ is a same-object drawing by a different artist. We apply the \textbf{triplet loss}
\begin{equation}
    \mathcal{L}_{\mathrm{tri}}
    =
    \max
    \left(
    0,
    d
    \left(
    \hat{\mathbf{z}}^{a},
    \hat{\mathbf{z}}^{p}
    \right)
    -
    d
    \left(
    \hat{\mathbf{z}}^{a},
    \hat{\mathbf{z}}^{n}
    \right)
    +
    m
    \right),
\label{eq:triplet}
\end{equation}
where $d$ is cosine distance and $m$ is the margin. We use batch-hard mining to select the hardest positive and negative for each anchor within a batch~\cite{hermans:17}.

The image encoder and projection head are jointly trained with
\begin{equation}
    \mathcal{L}
    =
    \mathcal{L}_{\mathrm{tri}}
    +
    \lambda
    \mathcal{L}_{\mathrm{beh}},
    \label{eq:total}
\end{equation}
where $\lambda$ balances content-aware metric learning and behavior-guided supervision.

\subsection{Verification Phase}
\label{sec:design-verification}
%
\paragraph{Content Suppression (\textit{M5}).}
Although \textit{M3} and \textit{M4} reduce the influence of drawing content during training, the learned embedding may still retain drawing content. Hence, at the verification phase, we suppress the content shared by drawings of the same object across training artists. We estimate this shared content as the mean embedding of each object and subtract it from every drawing of that object.

Let $x_{a',o}$ denote the drawing of object $o$ by artist $a'$, $\mathbf{u}(x_{a',o})$ its standardized image embedding, and $\mathcal{A}_{\mathrm{train}}$ the set of training artists. For each object $o$, we compute its corresponding object mean as
\begin{equation}
    \boldsymbol{\mu}_{o}
    =
    \frac{1}{|\mathcal{A}_{\mathrm{train}}|}
    \sum_{a'\in\mathcal{A}_{\mathrm{train}}}
    \mathbf{u}(x_{a',o}).
    \label{eq:content-mean}
\end{equation}

For every drawing $x_{a',o}$ used in verification, we subtract $\boldsymbol{\mu}_{o}$ from $\mathbf{u}(x_{a',o})$ and apply $\ell_2$ normalization to obtain its \textbf{content-suppressed representation $\mathbf{r}(x_{a',o})$}.

\paragraph{Reference Drawings Aggregation (\textit{M6}).}
To strengthen the artist identity shared across the reference drawings, we average the $N$ content-suppressed reference representations and apply $\ell_2$ normalization to obtain the \textbf{reference artist representation $\mathbf{t}_a$}.

\paragraph{Verification Distance (\textit{M7}).}
Our final goal is to determine whether the query drawing $q$ is created by the claimed artist $a$. Hence, we define the \textbf{verification distance $d(q,a)$} as the cosine distance between the content-suppressed query representation $\mathbf{r}(q)$ and the reference artist representation $\mathbf{t}_a$, produced by \textit{M5} and \textit{M6}, respectively.

\begin{table*}[!t]
\centering
\footnotesize
\setlength{\tabcolsep}{5pt}
\begin{tabular}{@{}ll cccccc c cccccc@{}}
\toprule
 & & \multicolumn{6}{c}{\textbf{AUC}~($\uparrow$)} & & \multicolumn{6}{c}{\textbf{EER}~($\downarrow$)} \\
\cmidrule(lr){3-8}\cmidrule(lr){10-15}
\textbf{Encoder} & \textbf{Method} & $N$=$1$ & $5$ & $10$ & $15$ & $20$ & $29$ & & $N$=$1$ & $5$ & $10$ & $15$ & $20$ & $29$ \\
\midrule
\multirow{2}{*}{DINOv2-B} & Baseline & 0.679 & 0.799 & 0.827 & 0.838 & 0.844 & 0.849 & & 0.368 & 0.274 & 0.250 & 0.241 & 0.235 & 0.231 \\
 & \textbf{\name{}} & \textbf{0.786} & \textbf{0.886} & \textbf{0.906} & \textbf{0.914} & \textbf{0.918} & \textbf{0.922} & & \textbf{0.288} & \textbf{0.197} & \textbf{0.176} & \textbf{0.168} & \textbf{0.163} & \textbf{0.160} \\
\midrule
\multirow{2}{*}{DINOv3} & Baseline & 0.664 & 0.780 & 0.809 & 0.820 & 0.826 & 0.832 & & 0.377 & 0.286 & 0.261 & 0.251 & 0.246 & 0.242 \\
 & \textbf{\name{}} & \textbf{0.773} & \textbf{0.862} & \textbf{0.881} & \textbf{0.889} & \textbf{0.892} & \textbf{0.896} & & \textbf{0.296} & \textbf{0.217} & \textbf{0.197} & \textbf{0.190} & \textbf{0.185} & \textbf{0.180} \\
\midrule
\multirow{2}{*}{CLIP} & Baseline & 0.626 & 0.745 & 0.778 & 0.792 & 0.799 & 0.806 & & 0.408 & 0.318 & 0.292 & 0.281 & 0.276 & 0.270 \\
 & \textbf{\name{}} & \textbf{0.738} & \textbf{0.843} & \textbf{0.868} & \textbf{0.878} & \textbf{0.884} & \textbf{0.890} & & \textbf{0.326} & \textbf{0.235} & \textbf{0.210} & \textbf{0.199} & \textbf{0.194} & \textbf{0.190} \\
\midrule
\multirow{2}{*}{MAE} & Baseline & 0.560 & 0.680 & 0.716 & 0.731 & 0.739 & 0.748 & & 0.459 & 0.372 & 0.344 & 0.333 & 0.326 & 0.318 \\
 & \textbf{\name{}} & \textbf{0.730} & \textbf{0.824} & \textbf{0.849} & \textbf{0.858} & \textbf{0.863} & \textbf{0.869} & & \textbf{0.332} & \textbf{0.251} & \textbf{0.228} & \textbf{0.220} & \textbf{0.216} & \textbf{0.210} \\
\midrule
\multirow{2}{*}{ResNet-50} & Baseline & 0.469 & 0.596 & 0.639 & 0.657 & 0.667 & 0.678 & & 0.522 & 0.428 & 0.395 & 0.380 & 0.371 & 0.363 \\
 & \textbf{\name{}} & \textbf{0.699} & \textbf{0.805} & \textbf{0.835} & \textbf{0.847} & \textbf{0.854} & \textbf{0.860} & & \textbf{0.356} & \textbf{0.268} & \textbf{0.240} & \textbf{0.229} & \textbf{0.222} & \textbf{0.216} \\
\bottomrule
\end{tabular}
\caption{\textbf{Overall verification performance across reference-set size $N$.} We compare the cross-entropy baseline with \name{} on five of the nine encoders. Each query drawing is compared with a reference set of $N$ drawings by the claimed artist. Values are means over 15 artist-disjoint splits and are computed on verification artists unseen during training. Results for all nine encoders are provided in the supplementary material.}
\label{tab:eval-overall}
\end{table*}

\begin{figure}[!t]
    \centering
    \includegraphics[width=\columnwidth]{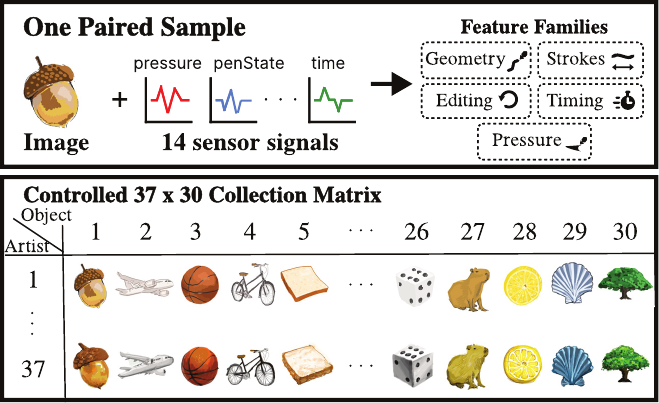}
    \caption{\textbf{Summary of the \name dataset.} Each cell of the $37{\times}30$
        matrix is a participant's drawing of the visual prompt, paired with the 14 tablet-pen sensor signals recorded during its creation.}
    \label{fig:data-collection}
\end{figure}
%
\section{Dataset and Experiment Setup}
\label{sec:protocol}
We construct a drawing--sensor dataset that links each completed drawing to the tablet-pen sensor signals recorded during its creation. As shown in Figure~\ref{fig:data-collection}, our dataset follows a fully crossed artist--object design. Each of 37 participants draws the same 30 objects, producing 1{,}110 drawing--sensor pairs. This design supports comparisons between different artists drawing the same object and between the same artist drawing different objects. We use this dataset for all experiments.

\paragraph{Data Collection.}
Participants use Apple Pencils on their own Apple iPads with an open-source tool~\cite{applepencil:web}. Each participant draws the same 30 objects using visual prompts, while the tool recorded 14 tablet-pen sensor signals, including geometry, strokes, editing, timing, and pressure. We conduct the study under our Institutional Review Board (IRB) approval. Further details on the participants, devices, drawing interface, object list, and collection procedure are provided in the supplementary material.

\paragraph{Evaluation Protocol.}
We form 15 artist-disjoint splits, each containing 23 fit, 6 validation, and 8 verification artists. In each split, all $8\times30=240$ verification drawings serve as queries. Each query drawing yields one genuine trial with $N$ different-object reference drawings by the same artist and seven impostor trials with same-object drawings by the other verification artists. This produces 240 genuine and 1{,}680 impostor trials per split. We vary the number of reference drawings as $N\in\{1,5,10,15,20,29\}$. For $N<29$, we average results over 20 random reference sets sampled from the artist's remaining objects. For $N=29$, all available reference drawings are used.

\paragraph{Metrics.}
For each comparison of a query drawing with the claimed artist's reference
drawings, we use cosine distance as the verification distance $d$. We sweep an acceptance threshold $\tau$ and accept the claim when $d\leq\tau$. We report the mean area under the receiver operating characteristic curve (AUC) and equal error rate (EER) over the 15 splits. AUC measures the overall separation between genuine and impostor trials without fixing a specific threshold. EER is the error rate at the threshold where the false acceptance rate equals the false rejection rate. We report both because AUC reflects overall performance, whereas EER reflects balanced error. We use a single seed and compare method variants with paired $t$-tests across splits. For results averaged across encoders, we also report the number of improved encoders. The supplementary material reports the false rejection rate (FRR) at a false acceptance rate (FAR) of $5\%$ and $10\%$, using thresholds selected on the validation artists.

\paragraph{Implementation.}
We fully fine-tune nine image encoders: DINOv2-S/B/L, DINOv3, CLIP,
SigLIP2, ConvNeXt, MAE, and ResNet-50. Each encoder uses a 256-dimensional projection head. Following prior image-based artist-classification work~\cite{tan:16, cetinic:18}, the cross-entropy baseline is trained to classify each drawing among the fit artists. For each encoder, we select the \textit{M3} behavior-loss weight $\lambda$ and training epoch by verification AUC on the validation artists. The triplet loss defined in \textit{M4} uses a cosine margin of $0.2$. All methods use the same verification protocol based on the fit artists. The baseline encoder learning rate is tuned separately, while the behavior loss and triplet loss methods share one learning rate per encoder. Full optimization settings are provided in the supplementary material. Each run uses a single NVIDIA RTX 4090.

\section{Evaluation}
\label{sec:eval}

\subsection{Overall Results}
\label{sec:eval-overall}
Table~\ref{tab:eval-overall} compares the cross-entropy baseline with \name across the reference-set sizes $N$. We report five representative encoders spanning four pretraining paradigms: self-distillation (DINOv2-B and DINOv3), language-image contrastive pretraining (CLIP), masked image modeling (MAE), and supervised classification (ResNet-50). Subsequent analyses use DINOv2-B when reporting results for a single encoder. The supplementary material provides complete results for all nine encoders for the overall comparison and ablations.

\noindent\textbf{(i) The cross-entropy baseline struggles to verify previously unseen artists.} The baseline is trained to classify the fit artists, whereas the verification artists are unseen during training. Its embedding distance is also strongly influenced by the drawing content rather than artist identity. Across the nine encoders, the baseline achieves an average AUC of $0.606$ at $N{=}1$. Its performance falls below chance on ResNet-50, with an AUC of $0.469$. These results show that classifying known artists does not transfer reliably to verifying previously unseen artists.

\noindent\textbf{(ii) \name improves every encoder at every $N$.}  This result holds for all nine encoders and all six reference-set sizes. Across all nine encoders and six reference-set sizes, all 54 combinations show significant improvements under paired $t$-tests across splits, with all $p$-values below $10^{-6}$. 
The largest gain occurs on ResNet-50 at $N{=}1$, where AUC improves by $48.8\%$ (from $0.469$ to $0.699$) and EER drops by $31.9\%$ (from $0.522$ to $0.356$) relative to the baseline. At $N{=}29$, the relative gain on ResNet-50 remains substantial, with AUC improving by $26.9\%$ (from $0.678$ to $0.860$) and EER dropping by $40.4\%$ (from $0.363$ to $0.216$).
The average gain across all nine encoders is larger when fewer reference drawings are available: $0.138$ AUC at $N{=}1$, compared with $0.103$ at $N{=}29$.

\begin{table}[!t]
\centering
\footnotesize
\setlength{\tabcolsep}{3pt}
\begin{tabular}{@{}ll cc r@{}}
\toprule
\textbf{Predicts} & \textbf{Space} & \textbf{AUC}~($\uparrow$) &
\textbf{EER}~($\downarrow$) & \textbf{$\Delta$AUC} \\
\midrule
--- (Frozen)        & ---        & 0.122 & 0.809 & --- \\
Absolute            & Raw        & 0.238 & 0.703 & $+0.116$ \\
Absolute            & Normalized & 0.473 & 0.518 & $+0.235$ \\
\textbf{Difference} & \textbf{Normalized} & \textbf{0.703} & \textbf{0.351} & $+0.230$ \\
\bottomrule
\end{tabular}
\caption{\textbf{Behavior loss ablation for \textit{M3} at $N{=}1$.} Each variant uses DINOv2-B and is trained only with the behavior loss. Values are means over the 15 artist-disjoint splits. Each successive design improves performance in all 15 splits ($p<10^{-9}$), and the same ordering holds for every $N$.}
\label{tab:eval-loss-design}
\end{table}
%
\begin{table}[!t]
\centering
\footnotesize
\setlength{\tabcolsep}{0.5pt}
\begin{tabular}{@{}l cc r c@{}}
\toprule
 & \textbf{AUC}$(\uparrow)$ & \textbf{EER}$(\downarrow)$ & \textbf{$\Delta$AUC} \\
\midrule
Baseline                                & 0.606 & 0.422 & --- \\
$+$ Behavior Loss (\textit{M3})                              & 0.654 & 0.387 & $+0.048$  \\
$+$ Triplet Loss (\textit{M4})                           & 0.717 & 0.340 & $+0.063$ & \\
\textbf{$+$ Content Suppress. (\textit{M5}) ($=$ \name{})} & \textbf{0.744} & \textbf{0.319} & $+0.027$ \\
\bottomrule
\end{tabular}
\caption{\textbf{Module ablation for \name at $N{=}1$.}  Results are averaged over nine encoders and 15 artist-disjoint splits. Starting from the cross-entropy baseline, each row cumulatively adds one module to the preceding row. $\Delta$AUC denotes the gain over the preceding row. }
\label{tab:eval-module}
\end{table}
%
\subsection{Ablation Studies}
\label{sec:eval-ablation}
%
\paragraph{Behavior Loss Ablation.}
Table~\ref{tab:eval-loss-design} compares four designs for using tablet-pen sensor signals with DINOv2-B. Frozen image embeddings achieve an AUC of only $0.122$ at $N{=}1$, far below chance. Predicting each drawing's absolute behavior feature values from the raw image features increases AUC to $0.238$. Applying the same supervision in the normalized image embedding space raises AUC further to $0.473$, although performance remains below chance. Together, these results indicate that behavior supervision is more effective in the normalized image embedding space, but predicting absolute behavior feature values remains insufficient for verification. Verification compares two drawings by their distance, so the final design instead supervises differences between drawings. For same-object drawings by different artists, it predicts differences in their standardized behavior features from differences in their normalized image embeddings. This directly guides the embedding differences used for verification with the corresponding behavior differences. This design is the only one that exceeds chance, reaching an AUC of $0.703$ and improving AUC by $0.230$ over absolute behavior prediction in the normalized image embedding space. This pairwise design is the behavior-guided supervision (\textit{M3}) used by \name  (Eq.~\eqref{eq:behavior}), and all subsequent experiments use it.

\paragraph{Module Ablation.}
We conduct the module ablation at $N{=}1$, the most challenging setting, where verification uses only one reference drawing. Table~\ref{tab:eval-module} cumulatively adds \textit{M3}, \textit{M4}, and \textit{M5} and reports the average results across the nine encoders. Behavior-guided supervision (\textit{M3})  with the behavior loss increases AUC by $0.048$ over the baseline and improves eight of the nine encoders. Seven of these improvements are significant under paired $t$-tests across splits ($p<0.05$), while DINOv3 shows only a small decrease of $0.004$. Content-aware metric learning (\textit{M4}) with the triplet loss adds another $0.063$ AUC, and content suppression (\textit{M5}) adds $0.027$. Both modules improve all nine encoders ($p\leq0.002$). The complete model with \textit{M3}, \textit{M4}, and \textit{M5} is \name. It outperforms the baseline on every encoder ($p<10^{-6}$), with a total AUC gain of $0.138$. It also reduces EER from $0.422$ to $0.319$. The encoder-level results show that \textit{M3} provides larger gains particularly when baseline verification performance is weaker. \textit{M3} improves AUC by $0.156$ on ResNet-50 and $0.066$ on SigLIP2, compared with $0.024$ on DINOv2-B and $0.014$ on CLIP.

\begin{figure}[!t]
    \centering
    \includegraphics[width=\columnwidth]{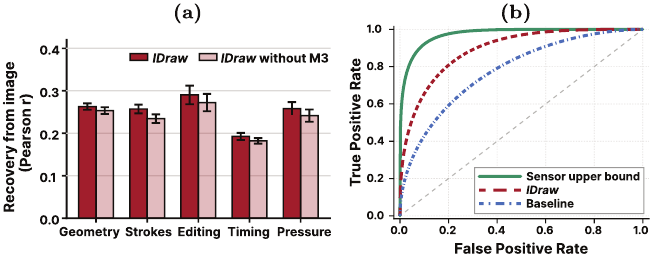}
    \caption{\textbf{Effect of sensor signals on DINOv2-B.}
    \textbf{(a)} Recovery of behavior feature differences from completed
    images, with and without behavior-guided supervision (\textit{M3}). Results show the mean $\pm$ one SEM over 15 splits.
    \textbf{(b)} Verification ROC curves under same-object impostors at $N{=}29$ on split~0. The model using ground-truth sensor signals provides an empirical sensor upper bound.}
    \label{fig:eval-sensor-signals}
\end{figure}
%
\subsection{Effect of Leveraging Sensor Signals}
\label{sec:eval-difficulty}
%
Figure~\ref{fig:eval-sensor-signals}(a) compares behavior feature prediction for full \name and \name without behavior-guided supervision (\textit{M3}) on DINOv2-B. We group the 50 behavior features into five families. Without behavior-guided supervision, prediction correlations range from approximately $r=0.18$ to $0.27$.  Behavior-guided supervision improves the correlation in all five families, with a mean gain of $+0.016$ across the 50 behavior features. Improvements occur in 12 of the 15 splits ($p{=}0.004$). At the individual behavior-feature level, the number of connected components in the strokes family improves by $+0.044$. This feature is directly reflected in the completed image. In contrast, the undo count in the revision family and total drawing time in the timing family are not directly observable, but may be reflected through overdraw and accumulated detail. Their correlations improve by $+0.033$ and $+0.031$, respectively. These small but consistent gains show that behavior-guided supervision makes drawing behavior more predictable from completed images.

Figure~\ref{fig:eval-sensor-signals}(b) estimates how much artist information the recorded sensor signals can provide. A model trained and evaluated directly on the ground-truth sensor signals reaches an AUC of $0.973$ at $N{=}29$. On DINOv2-B, \name closes $58.1\%$ of the gap between the baseline and this empirical sensor upper bound. The remaining gap may partly reflect behavior-related artist cues that are not fully recoverable from the completed image.

\section{Discussion and Conclusion}
\label{sec:conc}

\name shows that drawing behavior can support digital authorship verification even when sensor data are unavailable from the artist being verified. By learning from tablet-pen signals collected from separate training artists, \name extracts behavior-related artist cues from completed drawings, while content suppression reduces the influence of what the drawings depict. To enable this approach, we construct the first multimodal dataset for digital drawing authorship verification. Our design focuses on a practical setting in which reference drawings are reliably attributed to the claimed artist, as would be the case when they are enrolled through a trusted drawing application or provided directly by the artist. Extending verification to settings in which the provenance of the reference drawings is uncertain remains an important direction. Generative models introduce a further challenge because they can increasingly imitate an artist’s visual style. Whether they can also reproduce the behavior-related traces captured by \name remains an open question. However, we envision that \name can complement existing provenance mechanisms such as metadata and watermarking, providing an additional source of evidence when visual appearance alone is insufficient.

\bibliography{ref.main}

\end{document}